\documentclass[10pt, a4paper]{article}

\usepackage[final]{lrec2026} % this is the new style
\usepackage{tabularray}
\usepackage{comment}
\usepackage{parskip}

\title{Efficient Text Classification with Conformal In-Context Learning}

\name{Ippokratis Pantelidis\textsuperscript{1},
      Korbinian Randl\textsuperscript{1},
      Aron Henriksson\textsuperscript{1}}

\address{\textsuperscript{1}Stockholm University, Borgarfjordsgatan 12, 164 07 Kista, Sweden \\
         \{ippokratis.pantelidis, korbinian.randl, aronhen\}@dsv.su.se}

\abstract{
Large Language Models (LLMs) demonstrate strong in-context learning abilities, yet their effectiveness in text classification depends heavily on prompt design and incurs substantial computational cost. Conformal In-Context Learning (CICLe) has been proposed as a resource-efficient framework that integrates a lightweight base classifier with Conformal Prediction to guide LLM prompting by adaptively reducing the set of candidate classes. However, its broader applicability and efficiency benefits beyond a single domain have not yet been systematically explored. In this paper, we present a comprehensive evaluation of CICLe across diverse NLP classification benchmarks. The results show that CICLe consistently improves over its base classifier and outperforms few-shot prompting baselines when the sample size is sufficient for training the base classifier, and performs comparably in low data regimes. In terms of efficiency, CICLe reduces the number of shots and prompt length by up to 34.45\% and 25.16\%, respectively, and enables the use of smaller models with competitive performance. CICLe is furthermore particularly advantageous for text classification tasks with high class imbalance. These findings highlight CICLe as a practical and scalable approach for efficient text classification—combining the robustness of traditional classifiers with the adaptability of LLMs, achieving substantial gains in data and computational efficiency.
\\ \newline
}

\begin{document}

\maketitleabstract

\section{Introduction}

Text classification is a fundamental task in Natural Language Processing (NLP) with applications ranging from sentiment and emotion analysis to news categorization and knowledge organization. Despite its long history, the task remains challenging when datasets are characterized by a large number of classes, highly imbalanced class distributions, or limited training data. Recent advances in Large Language Models (LLMs) have introduced new possibilities through in-context learning \cite{brown2020language, chowdhery2022palm}, where models can adapt to unseen tasks from a small number of labeled examples, without further training. However, the effectiveness of few-shot prompting strongly depends on how examples are selected and presented \cite{liu2023lost}, and the computational cost of querying large models remains substantial.  

Conformal In-Context Learning (CICLe) was recently proposed as a resource-efficient approach that addresses these challenges \cite{randl2024cicle}. CICLe combines a traditional base classifier with Conformal Prediction (CP) to adaptively reduce the set of candidate classes and select corresponding few-shot examples to include in the prompt, where the LLM makes the final classification. This hybrid setup enables robust predictions while reducing the reliance on large context windows and bypassing the LLM entirely when the base classifier is sufficiently confident. Initial experiments on a food recall dataset~\cite{Randl2024_foodincidents} demonstrated that CICLe can outperform both traditional classifiers and regular few-shot prompting in terms of predictive performance and efficiency. However, its evaluation has so far been limited to a single domain, leaving open questions about CICLe’s generalizability and the conditions under which it provides the greatest benefit.

In this work, we conduct a systematic evaluation of CICLe across a diverse set of benchmark NLP datasets, including news categorization, ontology classification, community question answering, and fine-grained emotion detection. We also seek to understand in which kinds of situations CICLe performs well compared to its base classifier and regular few-shot prompting, focusing especially on efficiency aspects like sample size and model size, but also on the nature of the classification task, e.g. the number of classes and the class distribution. Our experiments build upon prior work in several ways:
\textbf{(i)}~we broaden the comparison between CICLe, its base classifier, and few-shot prompting by evaluating multiple selection strategies across diverse benchmark datasets;
\textbf{(ii)}~we analyze its robustness under varying training data sizes and LLM parameter scales; and
\textbf{(iii)}~we investigate its performance across class distributions ranging from balanced to highly imbalanced with class counts between 4 and 27. We frame our study around the following research question:  \textbf{under what conditions does CICLe provide advantages over traditional classification models and few-shot prompting strategies in text classification tasks with respect to sample size, the number of classes, and class imbalance.}

By answering this question, we aim to provide a comprehensive empirical understanding of CICLe’s strengths and limitations, and to identify when it represents a practical alternative for efficient and reliable text classification. In summary, our systematic evaluation across four benchmark datasets demonstrates that CICLe consistently improves over its base classifier and outperforms the few-shot baselines when there is a sufficient amount of data for training the base classifier. In low-data regimes, CICLe on average performs on par with tradiational few-shot prompting methods. In terms of efficiency, CICLe reduces the number of shots and the prompt length in tokens by up to 34.45\% and 25.16\%, respectively. The results also demonstrate that CICLe allows for using smaller models that perform competitively with much larger models, underscoring its potential for resource-efficient deployment. Furthermore, CICLe shows clear advantages in highly imbalanced scenarios, where it achieves stronger and more stable performance compared to regular few-shot prompting.

\section{Related Work}

\textbf{LLMs for few-shot classification} has reshaped the landscape of text classification by enabling models to perform tasks with only a handful of labeled examples provided in natural language prompts \cite{brown2020language}. This in-context learning paradigm bypasses the need for task-specific fine-tuning and has been applied successfully to a variety of NLP tasks \cite{gao2021making}. However, performance depends heavily on the choice and ordering of examples \cite{liu2023lost}, motivating strategies such as similarity-based retrieval \cite{rubin2022learning, ahmed2023improving} or embedding-based selection to improve reliability. Despite these advances, prompting remains sensitive to class imbalance, as LLMs may overpredict dominant classes while underpredicting others \cite{lin2024cobias}. While LLM prompting is powerful in low-data settings, several studies show that fine-tuning smaller models still achieves superior performance on many text classification tasks, suggesting that LLMs are not yet universally state-of-the-art for classification \cite{edwards2024language, bucher2024fine, alizadeh2023open}. On the other hand, it allows for carrying out classification tasks without having to first curate a large manually labeled dataset, which can be prohibitively expensive.

\textbf{Conformal prediction \citep[CP]{vovk2005algorithmic}} provides a statistical framework for associating predictions with calibrated confidence sets, offering formal guarantees of coverage. In essence, CP estimates how uncertain a model is by constructing a set of plausible labels for each prediction, such that the true label lies within this set with a predefined probability. This allows CP-based methods to quantify prediction reliability and control error rates in a principled way. Building on this idea, \citet{randl2024cicle} proposed \textbf{Conformal In-Context Learning (CICLe)}, which integrates CP with a base classifier to guide the selection of candidate classes for prompting. By adaptively reducing the number of classes passed to the LLM and bypassing the model when the base classifier is confident, CICLe addresses both reliability and efficiency. Initial results on food hazard classification showed that CICLe can outperform traditional classifiers and direct prompting while substantially lowering the computational cost. However, evaluation has so far been limited to a single dataset, leaving open questions about CICLe’s generalizability and the conditions under which it provides the greatest benefits.

As LLMs continue to grow in size and computational requirements, efficiency and accessibility in text classification have become increasingly important. Ideally, achieving strong classification performance should not depend solely on using the largest available models or extensive labeled datasets, as annotation remains costly and labor-intensive \cite{snow2008cheap, nie2019revealing}. Recent work has therefore emphasized approaches that balance effectiveness with practicality, aiming to reduce computational overhead while maintaining competitive results \cite{strubell2019energy, henderson2020towards, schick2021s, min2023recent}. However, most of these methods focus on efficient fine-tuning, or model compression rather than in-context classification, and are thus complementary to our evaluation. While \citet{purohit2025sampleefficientdemonstrationselection} explore few-shot classification, they focus on reinforcement learning of optimal sample combinations which remains computationally costly. By contrast, CICLe contributes to the growing effort toward efficient and sustainable NLP by combining a lightweight base classifier with selective and adaptive LLM prompting, thereby reducing dependence on massive models while maintaining strong performance.

\section{Methodology}

This section describes the methodological components of our study, comprising the TF-IDF–based base classifier, the few-shot prompting baselines used for comparison, and the CICLe framework, which extends the base classifier using conformal prediction for adaptive prompt construction.

\subsection{Base Classifier}
Following \citet{randl2024cicle}, we employ Logistic Regression trained on TF-IDF embeddings as our base classifier. This model provides a strong and efficient baseline against which we compare LLM-based methods. Notably, in the original CICLe study, this base classifier setup outperformed several BERT-family models, demonstrating that well-optimized traditional classifiers can remain highly competitive for text classification, particularly when computational efficiency is a priority. The base classifier also plays a central role in CICLe since its probability estimates are used for constructing conformal sets. While lightweight and competitive on well-represented classes, such linear models are known to degrade in low-data regimes and under strong class imbalance, motivating their integration with LLMs.  

\subsection{Few-Shot Prompting}
Few-shot prompting leverages LLMs by including a small number of labeled examples directly into the prompt, enabling the model to perform classification without fine-tuning and relying solely on in-context learning. It remains an open question how best to select examples for few-shot classification. In our experiments, we examine three complementary strategies for selecting these examples in order to establish strong few-shot baselines for comparison with CICLe. For all strategies, we use a total of $k=2$ labeled examples (shots) per class.

\textbf{Random selection (\emph{random})} serves as a simple baseline in which examples are sampled uniformly from the available training pool. This approach provides a neutral reference point for assessing the impact of more informed selection methods.

\textbf{Sparse similarity-based selection (\emph{sparse})} relies on representations derived from TF-IDF vectors. For each test instance, we compute cosine similarity to all training samples and select the most similar examples. This technique prioritizes examples with comparable lexical profiles to the input text, allowing the prompt to reflect characteristic word-level patterns of the target class.

\textbf{Dense similarity-based selection (\emph{dense})} operates in the semantic embedding space. Here, both training and test texts are encoded into dense vector representations using a pretrained sentence embedding model, and the closest examples are selected based on cosine similarity. Unlike the sparse approach, this method captures deeper semantic relationships beyond surface-level word overlap, often yielding more representative and contextually aligned examples.

\subsection{CICLe Framework}
CICLe integrates a base classifier with Conformal Prediction (CP) to create a resource-efficient prompting pipeline. The base classifier first estimates a probability distribution over the label space, which is then used to construct a conformal set of candidate classes that includes the true label with a predefined confidence level controlled by CP’s $\alpha$ parameter. In other words, $1 - \alpha$ represents the probability that the true class is contained in the conformal set. The LLM is prompted only with these candidate classes and with $k = 2$ few-shot examples from each, thereby reducing the context length and focusing the model on a smaller, more relevant subset of classes. In the limited case where the conformal set contains only one class, the system bypasses the LLM entirely and outputs the base classifier’s prediction directly. This hybrid mechanism not only reduces the computational cost of prompting but also improves robustness by adaptively balancing the strengths of traditional classifiers and LLMs.

\section{Experimental Setup}

This section details the experimental design used to evaluate CICLe and its baselines, including dataset selection, model configurations, prompt design, and implementation setup.

\subsection{Datasets}
\label{sec:datasets}
We evaluate our methods on four widely used text classification benchmarks that differ substantially in domain, text length, and label granularity: \emph{AG News}~\cite{Zhang2015_agnews} for topic categorization of news articles (4 classes), \emph{DBpedia-14}~\cite{Zhang2015_dbpedia} for ontology-based classification of Wikipedia abstracts (14 classes), \emph{Yahoo Answers Topics}~\cite{Zhang2015_yahooanswers} for community question categorization (10 classes), and \emph{GoEmotions}~\cite{Demszky2020_goemotions} for fine-grained emotion detection from short Reddit comments (27 classes). Since the original dataset allows multiple emotion labels per sample, we follow prior work in using the main (primary) emotion as the single target label. These datasets collectively allow us to assess CICLe under a variety of linguistic and structural conditions, including short-text, multiclass, and imbalanced scenarios.  

For each dataset, we experiment with varying numbers of available samples to simulate different data availability settings. Specifically, for training the base classifiers and constructing the few-shot example pool, we use subsets of 100, 200, 300, 400, 500, 1k, 2k, 3k, 4k, and 5k examples, employing identical random seeds to ensure consistency across models and runs. All single splits are stratified to maintain similar class distributions across training, calibration, and test sets. For the \emph{GoEmotions} dataset, which is highly imbalanced, we conduct experiments only on subsets of 2k samples and larger to ensure that all emotion categories are adequately represented. The evaluation is performed on a fixed test set of 1,000 examples for every dataset, providing a consistent and reproducible basis for comparison. A detailed overview of dataset statistics (average text length, class distribution, and total size) is provided in Table~\ref{tab:datasets}.

\begin{table*}[h!]
\centering
\begin{tabular}{lcccc}
\hline
\textbf{Dataset} & \textbf{Domain} & \textbf{Labels} & \textbf{Avg. length} & \textbf{Total size} \\
\hline
AG News & News articles & 4  & $\sim$40 tokens & 120,000 \\
DBpedia-14 & Wikipedia abstracts & 14 & $\sim$55 tokens & 560,000 \\
Yahoo Answers Topics & Community QA & 10 & $\sim$110 tokens & 1,400,000 \\
GoEmotions & Reddit comments & 27 & $\sim$13 tokens & 54,300 \\
\hline
\end{tabular}
\caption{Detailed statistics of the datasets used in our experiments.}
\label{tab:datasets}
\end{table*}

\subsection{Base Classifier}
The base classifier is a Logistic Regression model trained on TF-IDF embeddings with an $\ell_{2}$ penalty and a regularization strength of $C = 1.0$, corresponding to the default configuration in \texttt{scikit-learn}. This setup provides a simple and stable baseline while remaining fully consistent across datasets. The same configuration is used for the base classifier within the CICLe framework to ensure a fair and directly comparable evaluation. Unless otherwise specified, all experiments use $\alpha = 0.05$, corresponding to a 95\% confidence level for the conformal prediction step.

\subsection{Large Language Models}
We evaluate CICLe and few-shot prompting using models from the LLaMA family, specifically the \texttt{LLaMA-3.1-Instruct} variants with 8 billion and 70 billion parameters. These two configurations allow us to examine how model size influences performance and efficiency within the same architecture. The smaller 8B model represents a resource-friendly option suitable for most research environments, while the 70B model approximates the upper bound of commonly used open-weight models at the time of experimentation. While newer and larger open-weight LLMs (e.g., LLaMA-3.3 or Mixtral variants) have since been released, LLaMA-3.1 remains one of the best-performing and most widely benchmarked instruction-tuned models, making it a representative and well-established choice for evaluating the CICLe framework.

For generation, we use deterministic decoding without sampling and set \texttt{max\_new\_tokens = 5} to ensure that only a single label is produced per instance. Each prompt explicitly instructs the model to output only the final label, and no post-processing or filtering is applied. Predictions that do not match any valid class label are counted as incorrect, ensuring a strict and consistent evaluation across methods.

\subsection{Prompt Construction}
Both few-shot prompting and CICLe employ the same overall prompt structure: each prompt introduces the classification task, provides illustrative labeled examples, and finally includes the test instance to be classified. In all cases, the model is explicitly instructed to output only the final class label. The key difference lies in how class information is represented. In the few-shot setting, we include $k=2$ labeled examples for each class, thereby enumerating all classes through the provided examples. In contrast, CICLe implicitly communicates the most relevant classes by presenting labeled examples ordered from most to least probable according to the base classifier’s confidence scores. This adaptive ordering allows CICLe to focus the prompt on a smaller, contextually relevant subset of classes, thereby reducing the number of few-shot examples required. 

\subsection{Implementation Details}
All experiments were conducted on a server equipped with 8 NVIDIA RTX~A5500 GPUs, each with 24~GB of VRAM, allowing for efficient distribution of experiments across datasets and configurations. All models were implemented in Python using the \texttt{scikit-learn} library for traditional classifiers, the \texttt{crepes} library for CP, and the \texttt{transformers} framework for LLM-based experiments. Specifically, We use \texttt{scikit-learn}'s \texttt{TfidfVectorizer} for our \textit{sparse} embeddings and \texttt{all-MiniLM-L6-v2}~\cite{all-MiniLM-L6-v2} embedding model from HuggingFace's sentence transformers as our \textit{dense} embeddings.

Inference with the LLaMA-3.1 models was performed using the HuggingFace API without further fine-tuning. Random seeds were fixed across all runs to ensure full reproducibility. To promote transparency, the complete implementation, including code and configuration files, will be made publicly available on GitHub after the review phase and included in the camera-ready version.

\section{Results}
\label{sec:results}

In this section, we present the results of our experiments across four benchmark datasets. Unless otherwise noted, the primary evaluation metric is the macro-averaged F$_1$-score, which provides a balanced measure of performance across classes and is particularly informative under class imbalance. For each dataset, we plot the full learning curves for macro-F$_1$ across different training sample sizes, along with curves showing the reduction in the number of shots and candidate classes achieved by CICLe.

\subsection{Predictive Performance}

\begin{figure*}[h!]
    \centering
    \includegraphics[width=1.0\linewidth]{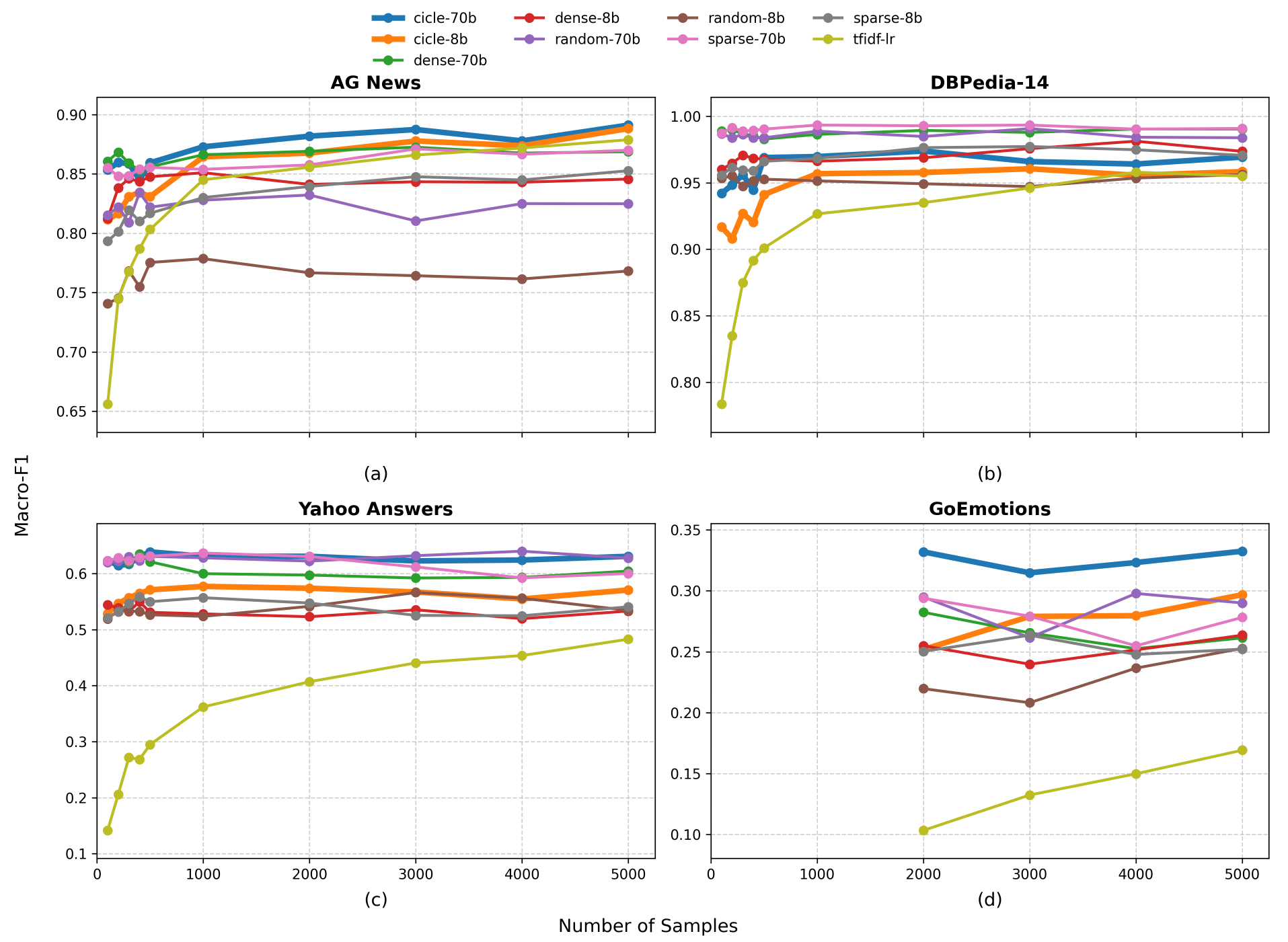}
    \caption[Macro F$_1$ vs. number of samples in the training set]{Macro F$_1$ vs number of samples in the training set. \emph{CICLe performs comparably to other strategies in balanced datasets and outperforms its competition in the imbalanced scenario.}}
    \label{fig:results_f1}
\end{figure*}

On the balanced \textbf{AG News} benchmark, CICLe achieves strong performance in terms of macro-F$_1$, confirming its effectiveness: Figure~\ref{fig:results_f1}a shows that CICLe outperforms other prompting strategies using the same LLM for training data sets with more than 500 labeled texts. With less than 500 samples, dense-70B slightly outperforms CICLe-70B, but CICLe quickly closes this gap as data increases. Notably, CICLe-8B not only surpasses all 8B few-shot variants but also performs comparably to larger models such as sparse-70B. Overall, we see very similar performance for all non-random tested strategies on this dataset.

On the balanced \textbf{DBpedia-14} dataset (see Figure~\ref{fig:results_f1}b), CICLe achieves consistently strong results across all sample sizes. Both CICLe variants outperform their base classifiers and random sample selection baselines, but perform minimally worse than dense or sparse strategies with the same LLM. Again we see largely comparable performance of the non-random prompting strategies.

The \textbf{Yahoo Answers Topics} dataset, while still being balanced, presents a more challenging, longer-text classification task with higher topical variability. Figure~\ref{fig:results_f1}c shows that for this dataset the LLM size makes a clear difference, as the strategies leveraging Llama-70B consistently outperform Llama-8B. CICLe consistently shows top performance for strategies based on the same LLM, and clearly outperforms its base classifier independent of the size of the training set.

\textbf{GoEmotions}, with its 27 imbalanced emotion categories, provides a challenging benchmark for fine-grained emotion classification. Here, CICLe-70B clearly achieves the highest macro-F$_1$ across all data sizes, while CICLe-8B remains competitive and often surpasses the 70B few-shot prompting variants. Overall, CICLe demonstrates stable and superior performance especially in this highly imbalanced setting as illustrated in Figure~\ref{fig:results_f1}d.

\subsection{Resource Efficiency}

\begin{figure*}[h!]
    \centering
    \includegraphics[width=1.0\linewidth]{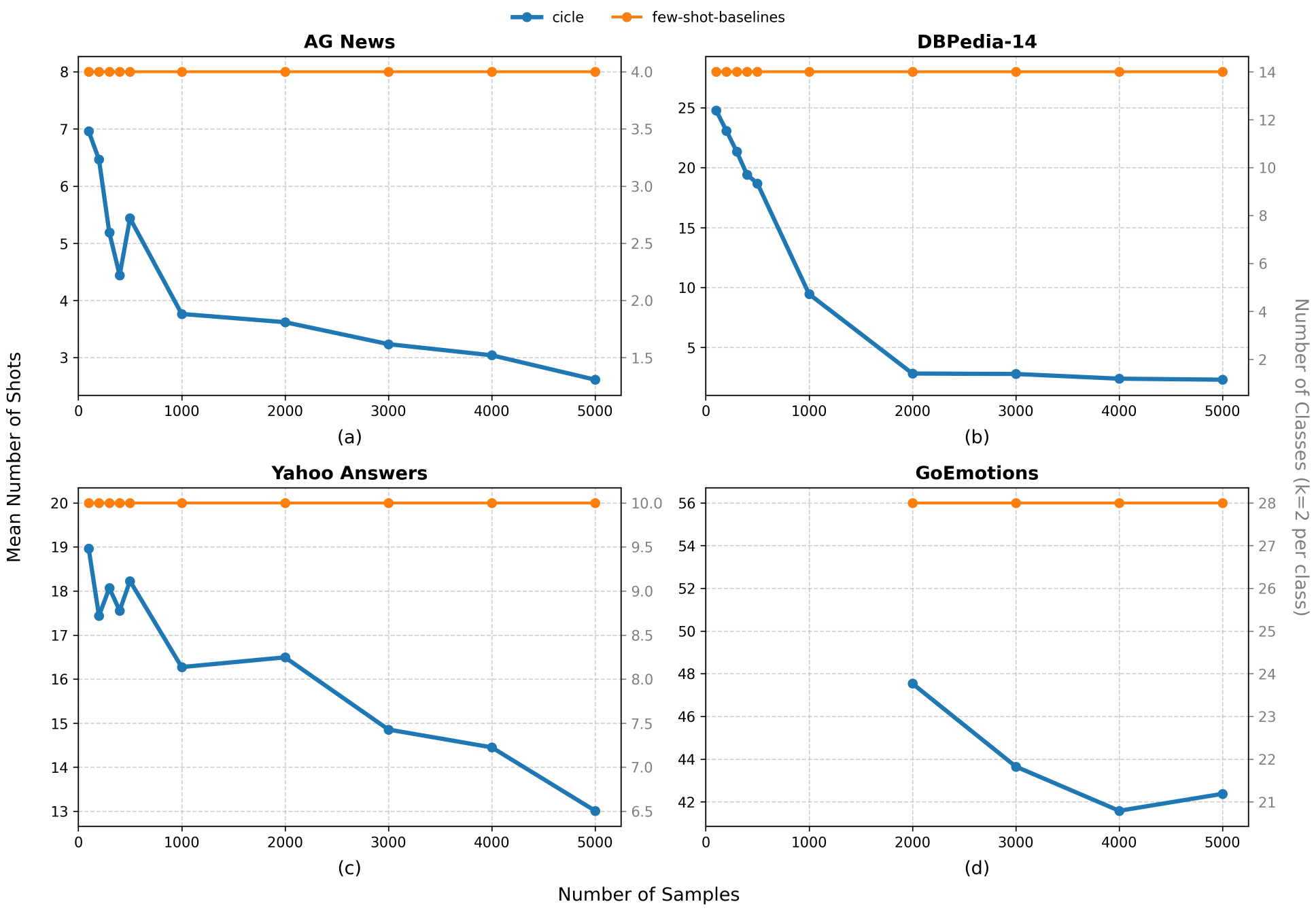}
    \caption[Prompt size vs. number of samples in the training set]{Prompt size vs. number of samples in the training set for CICLe vs. few-shot baselines (N.B. model size and shot selection strategy have no impact). \emph{CICLe consistently uses a lower number of shots in its prompts, with this number decreasing as the number of training samples increases and the base classifier gets better.}}
    \label{fig:results_shots}
\end{figure*}

While CICLe overall receives comparable or better F$_1$ scores than the baselines it needs consistently less few-shot samples to do so, as shown in Figure~\ref{fig:results_shots}. Even for small sizes of training sets, we see a reduction of the number of classes included in the prompt compared to prompting with the full set of classes for each dataset. Furthermore, we see the number of classes decreasing drastically with increasing size of the training set.

\subsection{General Trends}
Across datasets, several consistent trends emerge. First, CICLe shows stable performance across number of training samples that is either comparable (for balanced datasets) or (in the case of class imbalance) better than the tested dense or sparse few-shot selection strategies. Second, we find that CICLe is able to deliver its competitive performance with considerably fewer few-shot samples than its competition. Third,  on the highly imbalanced GoEmotions dataset, CICLe attains the best overall performance, indicating its effectiveness in handling uneven label distributions. Finally, CICLe also reliably outperforms its base classifier, reflecting the additional reasoning capabilities contributed by the LLM.

Regarding few-shot prompting, dense and sparse strategies perform overall comparably, with the sparse strategy often slightly outperforming dense few-shot selection. Random selection remains consistently unstable. Overall, CICLe combines the efficiency of classical models with the adaptability of LLMs, achieving robust and resource-efficient text classification across diverse linguistic and structural conditions.

\subsection{Metrics Across Different Datasets}
To provide an overall comparison of model performance across datasets, we report the average macro-F$_1$ scores of each method computed over all sample sizes for each dataset separately in Table~\ref{tab:aggregate_results}. This aggregation captures the general behavior of each model while smoothing out small fluctuations caused by random sampling or data scarcity.

Both CICLe strategies achieve the best performance overall compared to sparse and dense strategies using the same size LLM, ranking first on three of the four datasets (AG News, Yahoo Answers, and GoEmotions) for the 8B variant and two out of four (AG News, GoEmotions) for the 70B variant. The smaller CICLe-8B model remains highly competitive, sometimes even closely matching several 70B prompting variants. Among the few-shot baselines, we again see no clear winner between dense and sparse strategies. The base classifier remains strong on AG News and DBpedia but drops substantially on the more complex and imbalanced datasets, highlighting the role of CP in CICLe’s steady performance in those settings.

\begin{table}[htbp]
\centering
\small
\setlength{\tabcolsep}{4pt} 
\resizebox{\linewidth}{!}{
\begin{tabular}{l@{\hskip 4pt}cccc}
\hline
    \textbf{Model} & \textbf{AG} & \textbf{DBp.} & \textbf{Yahoo} & \textbf{GoEm.} \\
\hline

    \vspace{5pt}
    Base Classifier        & 0.808          & 0.901          & 0.333          & 0.139 \\

    Few-shot (Rnd., 8B)    & 0.762          & 0.952          & 0.537          & 0.229 \\
    Few-shot (Sprs., 8B)   & 0.823          & 0.967          & 0.540          & 0.253 \\
    Few-shot (Dens., 8B)   & 0.841          & \textbf{0.970} & 0.533          & 0.252 \\
    \vspace{5pt}
    CICLe (8B)             & \textbf{0.850} & 0.940          & \textbf{0.561} & \textbf{0.277} \\

    Few-shot (Rnd., 70B)   & 0.822          & 0.986          & \textbf{0.627} & 0.286 \\
    Few-shot (Sprs., 70B)  & 0.858          & \textbf{0.991} & 0.620          & 0.277 \\
    Few-shot (Dens., 70B)  & 0.864          & 0.988          & 0.611          & 0.265 \\
    CICLe (70B)            & \textbf{0.869} & 0.960          & 0.626          & \textbf{0.326} \\
\hline  
\end{tabular}
}
\caption{Average macro-F$_1$ scores across all sample sizes for each model and dataset. The best result per dataset is in bold.}
\label{tab:aggregate_results}
\end{table}

\subsection{Metrics for Different Data Regimes}
To further examine CICLe’s data efficiency, we analyze performance across three data availability regimes by aggregating results over comparable sample sizes. Specifically, we categorize training set sizes into three groups: \textit{low-data} (100–400 samples), \textit{medium-data} (500–2,000 samples), and \textit{large-data} (3,000–5,000 samples). For each group, we compute average macro-F$_1$ scores across the AG News and Yahoo Answers datasets, enabling us to study how each method scales with increasing amounts of training data across multiple datasets. 

We exclude DBpedia-14 from this analysis due to indications of potential data contamination, reflected in unusually high F$_1$ scores for the sparse and dense few-shot baselines. Since DBpedia-14 consists of Wikipedia abstracts, which overlap with the pretraining data of many large language models, this dataset may not provide a fully independent evaluation of in-context classification performance. Similarly, we exclude GoEmotions, as it is evaluated only for sample sizes above 2k (see Section~\ref{sec:datasets}) to ensure adequate class representation. This selection ensures direct comparability across regimes, with all remaining datasets contributing uniformly.

\begin{table}[htbp]
\centering
\small
\setlength{\tabcolsep}{4pt}
\begin{tabular}{l@{\hskip 4pt}ccc}
\hline
\textbf{Model} & \textbf{Low} & \textbf{Medium} & \textbf{Large} \\
\hline
    \vspace{5pt}
    Base Classifier        & 0.4803 & 0.5946 & 0.6656 \\

    Few-shot (Rnd., 8B)    & 0.6408 & 0.6520 & 0.6584 \\
    Few-shot (Sprs., 8B)   & 0.6726 & 0.6900 & 0.6893 \\
    Few-shot (Dens., 8B)   & \textbf{0.6881} & 0.6869 & 0.6866 \\
    \vspace{5pt}
    CICLe (8B)             & 0.6862 & \textbf{0.7140} & \textbf{0.7220} \\

    Few-shot (Rnd., 70B)   & 0.7219 & 0.7271 & 0.7265 \\
    Few-shot (Sprs., 70B)  & 0.7385 & 0.7441 & 0.7352 \\
    Few-shot (Dens., 70B)  & \textbf{0.7430} & 0.7348 & 0.7330 \\
    CICLe (70B)            & 0.7376 & \textbf{0.7524} & \textbf{0.7555} \\
\hline  
\end{tabular}
\caption{Average macro-F$_1$ scores across all sample sizes for each model and dataset. CICLe-70B achieves the highest performance across two datasets (AG News and Yahoo), demonstrating its strong and consistent advantage over both few-shot prompting and base classifier baselines. The best result per dataset is shown in bold.}
\label{tab:sample_size_results}
\end{table}

Table~\ref{tab:sample_size_results} reports the resulting averages, illustrating how CICLe and few-shot prompting methods behave under different data regimes.
CICLe-70B achieves the highest overall macro-F$_1$ scores across data regimes, ranking first in the medium and large settings and and achieving comparably to top performances in low-data conditions. The smaller CICLe-8B model also performs best among the 8B variants, matching or slightly surpassing the larger few-shot models as the amount of training data increases. These findings confirm that CICLe maintains strong generalization and delivers substantial efficiency gains, allowing smaller models to approach the performance of much larger LLMs. While most methods improve with more data, the gains are not uniform: for example, few-shot dense prompting with the 70B model slightly decreases in the large-data regime, suggesting diminishing returns from additional examples. The base classifier also benefits from more data but continues to lag behind CICLe, particularly in low-data scenarios where conformal calibration and LLM reasoning provide the greatest advantage. Overall, CICLe demonstrates robustness and scalability, maintaining competitive performance while requiring fewer computational resources.

\subsection{Prompt \& Shots Reduction}

To further assess the efficiency gains of CICLe, we compare the average prompt length and number of few-shot examples (\emph{shots}) used per prediction against standard few-shot prompting. 
The percentage reductions are computed relative to the average of the three few-shot baselines (\emph{dense}, \emph{sparse}, and \emph{random}) across all sample sizes. As shown in Table~\ref{tab:prompt_reduction}, CICLe consistently requires fewer examples, usually resulting in shorter prompts, and highlighting its efficiency in resource-constrained settings. The only exception is the Yahoo dataset, where prompt length slightly increases due to the sparse baseline selecting generally shorter texts. Nevertheless, the number of shots still decreases, confirming CICLe’s ability to reduce context size even in this case.

\begin{table}[htbp]
\centering
\small
\setlength{\tabcolsep}{4pt}
\begin{tabular}{lcccc}
\hline
\textbf{Avg. Reduction (\%)} & \textbf{AG} & \textbf{DBp.} & \textbf{Yahoo} & \textbf{GoEm.} \\
\hline
Prompt Length & 34.73 & 49.91 & -2.51 & 18.93 \\
Num. Shots    & 44.04 & 54.62 & 17.33 & 21.81 \\
\hline
\end{tabular}
\caption{Average reduction in prompt length and number of few-shot examples per prediction for CICLe compared to the mean of the three few-shot baselines (\emph{dense}, \emph{sparse}, and \emph{random}).}
\label{tab:prompt_reduction}
\end{table}

\section{Discussion \& Conclusion}
\label{sec:discussion}

This study set out to examine under what conditions CICLe provides advantages over traditional classifiers and few-shot prompting strategies in text classification. Our systematic evaluation across four diverse benchmarks shows that CICLe consistently improves over its base classifier and performs competitively with few-shot prompting methods. The results across datasets highlight several consistent insights about the behavior and applicability of CICLe. First, our experiments on a diverse set of datasets and tasks largely support the findings of \citet{randl2024cicle}: we find that CICLe performs on par or better than other few-shot selection approaches at substantially shorter prompt length, up to 50\% in some cases. This reduced prompt length leads to lower runtime and memory requirements because of the causal masking mechanism in transformer models and by extension LLMs: although these models have a fixed context length, attention weights are only computed for tokens in the input prompt \citep{vasvani2017transformer, touvron2023llama}.

Second, as also shown by \citet{randl2024cicle}, the framework provides a robust and data-efficient improvement over its base classifier. This advantage is particularly evident in imbalanced settings, where CICLe maintains reliable performance by effectively combining the strengths of lightweight traditional models and LLM-based reasoning. It thereby bridges the gap between traditional classifiers and large-scale prompting methods, achieving competitive results without requiring extensive fine-tuning or large labeled datasets.

Nevertheless, our findings also include previously unpublished insights. In very low-data regimes (less than 500 training samples), CICLe and few-shot prompting exhibit comparable performance, with minor differences likely arising from how examples are selected and ordered. In these cases, the base classifier’s limited confidence only leads to a small reduction of the candidate label set, resulting in a similar number of shots to standard few-shot prompting. Consequently, performance mainly depends on the quality and ordering of the examples rather than on the framework itself. Although CICLe relies on sparse embeddings for similarity estimation, the few-shot sparse and dense baselines occasionally show marginal variation, possibly influenced by the different ordering of examples. This aspect is not systematically evaluated in this work and left for future investigation. Importantly, there is no theoretical reason for CICLe to perform worse than regular few-shot prompting, as both rely on the same underlying model behavior and differ only in how examples and candidate labels are selected.

Furthermore, our experiments reveal that CICLe often outperforms other strategies for training datasets with more than 500 samples. This is especially true for imbalanced datasets, where CICLe's ability to balance between the computationally cheap but weaker base classifier and the computationally costly but more powerful LLM ensures optimal handling of each class independently of its support. Beyond these observations, our results confirm that smaller instruction-tuned models can reach nearly the same performance as much larger ones when used within the CICLe framework, reinforcing its value as a practical and sustainable approach to efficient text classification.

Future work will focus on a deeper examination of the $\alpha$ parameter, which controls the confidence level in CICLe and thus governs the trade-off between prediction certainty and label coverage. Another important direction is to more thoroughly experiment with CICLe using dense embeddings, as well as to study the impact of example ordering in the prompt, while may also influence performance. Finally, we plan to extend CICLe to more complex tasks such as multi-label or hierarchical classification, where managing uncertainty across overlapping categories remains a key challenge.

Overall, CICLe provides a simple yet powerful framework that combines the efficiency of traditional classifiers with the adaptability of LLMs for reliable text classification. It is particularly advantageous in scenarios that require a balance between performance and efficiency, offering strong results with shorter prompts, fewer examples, and smaller LLMs. By demonstrating when and how CICLe is most beneficial, this work establishes a solid empirical foundation for applying CICLe in resource-conscious NLP applications.

\section*{Acknowledgments}
\textit{Acknowlegdements will be added after the end of the anonymity period.}
This research has been funded by the European Union’s Horizon Europe research and innovation program EFRA (Grant Agreement Number 101093026). Views and opinions expressed are however those of the authors only and do not necessarily reflect those of the European Union or European Commission-EU. Neither the European Union nor the granting authority can be held responsible for them. {\normalsize\euflag}

\section*{Limitations}
One limitation relates to model diversity. Our evaluation focuses solely on two parameter variations of the LLaMA-3.1-Instruct model, selected to study efficiency and performance trade-offs within a single architecture family. While this provides a controlled comparison, it does not capture the full range of behaviors that might arise with other LLM architectures or instruction-tuned variants.

Finally, all experiments were conducted exclusively on English datasets. Although this choice aligns with prior work on CICLe and facilitates comparability, it leaves open questions about the framework’s applicability to multilingual, or even domain-specific text classification tasks such as those in medicine, law, or other specialized fields.

\section*{Bibliographical References}

\section*{Language Resource References}


@LanguageResource{Zhang2015_agnews,
 title     = {AG News},
 author    = {Xiang Zhang},
 url       = {https://huggingface.co/datasets/fancyzhx/ag\_news},
 pid       = {https://huggingface.co/datasets/fancyzhx/ ag\_news},
 booktitle = {NIPS},
 publisher = {HuggingFace.co},
 year      = {2015}
}

@LanguageResource{Zhang2015_dbpedia,
 title     = {DBpedia14},
 author    = {Xiang Zhang},
 url       = {https://huggingface.co/datasets/fancyzhx/dbpedia\_14},
 pid       = {https://huggingface.co/datasets/ fancyzhx/dbpedia\_14},
 booktitle = {NIPS},
 publisher = {HuggingFace.co},
 year      = {2015}
}

@LanguageResource{Zhang2015_yahooanswers,
 title     = {Yahoo Answers Topics},
 author    = {Xiang Zhang},
 url       = {https://huggingface.co/datasets/community-datasets/yahoo\_answers\_topics},
 pid       = {https://huggingface.co/datasets/community-datasets/yahoo\_answers\_topics},
 booktitle = {NIPS},
 publisher = {HuggingFace.co},
 year      = {2015}
}

@LanguageResource{Demszky2020_goemotions,
 title     = {{GoEmotions: A Dataset of Fine-Grained Emotions}},
 author    = {Demszky, Dorottya and Movshovitz-Attias, Dana and Ko, Jeongwoo and Cowen, Alan and Nemade, Gaurav and Ravi, Sujith},
 url       = {https://huggingface.co/datasets/google-research-datasets/go\_emotions},
 pid       = {https://huggingface.co/datasets/google-research-datasets/go\_emotions},
 booktitle = {58th Annual Meeting of the Association for Computational Linguistics (ACL)},
 publisher = {HuggingFace.co},
 year      = {2020}
}

@LanguageResource{Randl2024_foodincidents,
  author       = {Randl, Korbinian and
                  Karvounis, Manos and
                  Marinos, George and
                  Pavlopoulos, John and
                  Lindgren, Tony and
                  Henriksson, Aron},
  title        = {Food Recall Incidents},
  year         = 2024,
  publisher    = {Zenodo},
  pid          = {https://doi.org/10.5281/zenodo.10891602},
  url          = {https://doi.org/10.5281/zenodo.10891602},
}

@LanguageResource{all-MiniLM-L6-v2,
  author       = {Nils Reimers and Tom Arsen and others},
  title        = {sentence-transformers/all-MiniLM-L6-v2},
  year         = 2024,
  publisher    = {HuggingFace.co},
  pid          = {https://huggingface.co/sentence-transformers/all-MiniLM-L6-v2},
  url          = {https://huggingface.co/sentence-transformers/all-MiniLM-L6-v2},
}

@inproceedings{Zhang2015CharacterlevelCN,
  title={Character-level Convolutional Networks for Text Classification},
  author={Xiang Zhang and Junbo Jake Zhao and Yann LeCun},
  booktitle={NIPS},
  year={2015}
}

\begin{thebibliography}{00}

\bibitem[\protect\citeauthoryear{Brown et al.}{Brown et al.}{2020}]{brown2020language}
Brown, T., Mann, B., Ryder, N., Subbiah, M., Kaplan, J., Dhariwal, P., \ldots \& others.
(2020).
Language models are few-shot learners.
\emph{Advances in Neural Information Processing Systems}, 33, 1877--1901.

\bibitem[\protect\citeauthoryear{Chowdhery et al.}{Chowdhery et al.}{2022}]{chowdhery2022palm}
Chowdhery, A., Narang, S., Devlin, J., Bosma, M., Mishra, G., Roberts, A., \ldots \& others.
(2022).
PaLM: Scaling language modeling with pathways.
\emph{arXiv preprint arXiv:2204.02311}.

\bibitem[\protect\citeauthoryear{Liu et al.}{Liu et al.}{2023}]{liu2023lost}
Liu, N. F., Lin, K., Hewitt, J., Paranjape, A., Bevilacqua, M., Petroni, F., \& Liang, P.
(2023).
Lost in the Middle: How Language Models Use Long Contexts.
\emph{arXiv preprint arXiv:2307.03172}.

\bibitem[\protect\citeauthoryear{Randl et al.}{Randl et al.}{2024}]{randl2024cicle}
Randl, K., Pavlopoulos, J., Henriksson, A., \& Lindgren, T.
(2024).
CICLe: Conformal In-Context Learning for Large-scale Multi-Class Food Risk Classification.
In \emph{Findings of the Association for Computational Linguistics: ACL 2024} (pp. 7695--7715).
Bangkok, Thailand: Association for Computational Linguistics.

\bibitem[\protect\citeauthoryear{Edwards \& Camacho-Collados}{Edwards \& Camacho-Collados}{2024}]{edwards2024language}
Edwards, A., \& Camacho-Collados, J.
(2024).
Language Models for Text Classification: Is In-Context Learning Enough?
\emph{arXiv preprint arXiv:2403.17661}.

\bibitem[\protect\citeauthoryear{Bucher \& Martini}{Bucher \& Martini}{2024}]{bucher2024fine}
Bucher, L., \& Martini, R.
(2024).
Fine-Tuned 'Small' LLMs (Still) Significantly Outperform Zero-Shot Generative AI Models in Text Classification.
\emph{arXiv preprint arXiv:2406.08660}.

\bibitem[\protect\citeauthoryear{Alizadeh et al.}{Alizadeh et al.}{2023}]{alizadeh2023open}
Alizadeh, M., Jahanbakhsh, K., Darvishi, E., \& Veisi, H.
(2023).
Open-Source LLMs for Text Annotation: A Practical Guide.
\emph{arXiv preprint arXiv:2307.02179}.

\bibitem[\protect\citeauthoryear{Gao et al.}{Gao et al.}{2021}]{gao2021making}
Gao, T., Fisch, A., \& Chen, D.
(2021).
Making pre-trained language models better few-shot learners.
\emph{arXiv preprint arXiv:2012.15723}.

\bibitem[\protect\citeauthoryear{Rubin et al.}{Rubin et al.}{2022}]{rubin2022learning}
Rubin, O., Herzig, J., \& Berant, J.
(2022).
Learning to retrieve prompts for in-context learning.
\emph{arXiv preprint arXiv:2112.08633}.

\bibitem[\protect\citeauthoryear{Ahmed et al.}{Ahmed et al.}{2023}]{ahmed2023improving}
Ahmed, T., Pai, K. S., Devanbu, P., \& Barr, E. T.
(2023).
Improving few-shot prompts with relevant static analysis products.
\emph{arXiv preprint arXiv:2304.06815}.

\bibitem[\protect\citeauthoryear{Vovk et al.}{Vovk et al.}{2005}]{vovk2005algorithmic}
Vovk, V., Gammerman, A., \& Shafer, G.
(2005).
\emph{Algorithmic learning in a random world}.
Springer.

\bibitem[\protect\citeauthoryear{Touvron et al.}{Touvron et al.}{2023}]{touvron2023llama}
Touvron, H., Martin, L., Stone, K., \& others.
(2023).
LLaMA 2: Open foundation and fine-tuned chat models.
\emph{arXiv preprint arXiv:2307.09288}.

\bibitem[\protect\citeauthoryear{Schick \& Schütze}{Schick \& Schütze}{2021}]{schick2021s}
Schick, T., \& Schütze, H.
(2021).
It's not just size that matters: Small language models are also few-shot learners.
In \emph{Proceedings of the 2021 Conference of the North American Chapter of the Association for Computational Linguistics} (pp. 2339--2352).

\bibitem[\protect\citeauthoryear{Henderson et al.}{Henderson et al.}{2020}]{henderson2020towards}
Henderson, P., Hu, J., Romoff, J., Brunskill, E., Jurafsky, D., \& Pineau, J.
(2020).
Towards the systematic reporting of the energy and carbon footprints of machine learning.
\emph{Journal of Machine Learning Research}, 21(248), 1--43.

\bibitem[\protect\citeauthoryear{Strubell et al.}{Strubell et al.}{2019}]{strubell2019energy}
Strubell, E., Ganesh, A., \& McCallum, A.
(2019).
Energy and policy considerations for deep learning in NLP.
In \emph{Proceedings of the 57th Annual Meeting of the Association for Computational Linguistics} (pp. 3645--3650).

\bibitem[\protect\citeauthoryear{Snow et al.}{Snow et al.}{2008}]{snow2008cheap}
Snow, R., O'Connor, B., Jurafsky, D., \& Ng, A. Y.
(2008).
Cheap and fast---but is it good? Evaluating non-expert annotations for natural language tasks.
In \emph{Proceedings of the 2008 Conference on Empirical Methods in Natural Language Processing} (pp. 254--263).

\bibitem[\protect\citeauthoryear{Nie et al.}{Nie et al.}{2019}]{nie2019revealing}
Nie, A., Bennett, M., \& Goodman, N.
(2019).
Revealing the importance of semantic retrieval for machine reading at scale.
In \emph{Proceedings of the 2019 Conference on Empirical Methods in Natural Language Processing} (pp. 2553--2564).

\bibitem[\protect\citeauthoryear{Min et al.}{Min et al.}{2023}]{min2023recent}
Min, S., Lewis, M., Zettlemoyer, L., \& Hajishirzi, H.
(2023).
Recent advances in efficient fine-tuning of transformer-based models.
In \emph{Proceedings of the 61st Annual Meeting of the Association for Computational Linguistics}.

\bibitem[\protect\citeauthoryear{Lin \& You}{Lin \& You}{2024}]{lin2024cobias}
Lin, R., \& You, Y.
(2024).
The Curious Case of Class Accuracy Imbalance in LLMs: Post-hoc Debiasing via Nonlinear Integer Programming.
\emph{arXiv preprint arXiv:2405.07623}.

\bibitem[\protect\citeauthoryear{Randl et al.}{Randl et al.}{2025}]{randl2025foodhazard}
Randl, K., Pavlopoulos, J., Henriksson, A., Lindgren, T., \& Bakagianni, J.
(2025).
SemEval-2025 Task 9: The Food Hazard Detection Challenge.
In \emph{Proceedings of the 19th International Workshop on Semantic Evaluation (SemEval-2025)} (pp. 2523--2534).
Vienna, Austria: Association for Computational Linguistics.

\bibitem[\protect\citeauthoryear{Vaswani et al.}{Vaswani et al.}{2017}]{vasvani2017transformer}
Vaswani, A., Shazeer, N., Parmar, N., Uszkoreit, J., Jones, L., Gomez, A. N., Kaiser, \L{}, \& Polosukhin, I.
(2017).
Attention is all you need.
In \emph{Proceedings of the 31st International Conference on Neural Information Processing Systems} (pp. 6000--6010).

\bibitem[\protect\citeauthoryear{Purohit et al.}{Purohit et al.}{2025}]{purohit2025sampleefficientdemonstrationselection}
Purohit, K., Venktesh, V., Bhattacharya, S., \& Anand, A.
(2025).
Sample Efficient Demonstration Selection for In-Context Learning.
\emph{arXiv preprint arXiv:2506.08607}.

\end{thebibliography}

\begin{thebibliography}{00}

\bibitem[\protect\citeauthoryear{Zhang}{Zhang}{2015}]{Zhang2015_agnews}
Zhang, X. (2015).
\textit{AG News}.
HuggingFace.co.
\url{https://huggingface.co/datasets/fancyzhx/ag\_news}
(PID: \url{https://huggingface.co/datasets/fancyzhx/ag\_news}).

\bibitem[\protect\citeauthoryear{Zhang}{Zhang}{2015}]{Zhang2015_dbpedia}
Zhang, X. (2015).
\textit{DBpedia14}.
HuggingFace.co.
\url{https://huggingface.co/datasets/fancyzhx/dbpedia\_14}
(PID: \url{https://huggingface.co/datasets/fancyzhx/dbpedia\_14}).

\bibitem[\protect\citeauthoryear{Zhang}{Zhang}{2015}]{Zhang2015_yahooanswers}
Zhang, X. (2015).
\textit{Yahoo Answers Topics}.
HuggingFace.co.
\url{https://huggingface.co/datasets/community-datasets/yahoo\_answers\_topics}
(PID: \url{https://huggingface.co/datasets/community-datasets/yahoo\_answers\_topics}).

\bibitem[\protect\citeauthoryear{Demszky et al.}{Demszky et al.}{2020}]{Demszky2020_goemotions}
Demszky, D., Movshovitz-Attias, D., Ko, J., Cowen, A., Nemade, G., \& Ravi, S. (2020).
\textit{GoEmotions: A Dataset of Fine-Grained Emotions}.
In \emph{58th Annual Meeting of the Association for Computational Linguistics (ACL)}.
HuggingFace.co.
\url{https://huggingface.co/datasets/google-research-datasets/go\_emotions}
(PID: \url{https://huggingface.co/datasets/google-research-datasets/go\_emotions}).

\bibitem[\protect\citeauthoryear{Randl et al.}{Randl et al.}{2024}]{Randl2024_foodincidents}
Randl, K., Karvounis, M., Marinos, G., Pavlopoulos, J., Lindgren, T., \& Henriksson, A. (2024).
\textit{Food Recall Incidents}.
Zenodo.
\url{https://doi.org/10.5281/zenodo.10891602}
(PID: \url{https://doi.org/10.5281/zenodo.10891602}).

\bibitem[\protect\citeauthoryear{Reimers et al.}{Reimers et al.}{2024}]{all-MiniLM-L6-v2}
Reimers, N., Arsen, T., \& others. (2024).
\textit{sentence-transformers/all-MiniLM-L6-v2}.
HuggingFace.co.
\url{https://huggingface.co/sentence-transformers/all-MiniLM-L6-v2}
(PID: \url{https://huggingface.co/sentence-transformers/all-MiniLM-L6-v2}).

\end{thebibliography}
\end{document}